\documentstyle[eaclap]{article}

\title{\vspace{-0.5in}On Learning More Appropriate Selectional
Restrictions}

\author{Francesc Ribas\thanks{Revised version prepared for the CMP-LG
E-Print Archive of the original paper published in the Proceedings of the 7th
Conference of the European Chapter of the Association for
Computational Llinguistics, Dublin, Ireland, March 1995. The research
reported here has been made in the framework of the Acquilex-II Esprit
Project (7315), and has been supported by a grant of Departament
d'Ensenyament, Generalitat de Catalunya, 91-DOGC-1491.}¯
\\ Departament de Llenguatges i Sistemes Inform\`{a}tics
\\ Universitat Polit\`{e}cnica de Catalunya
\\ Pau Gargallo, 5
\\ 08028 Barcelona
\\ Spain
\\ {\tt ribas@lsi.upc.es}}

\begin{document}
\maketitle
\vspace{-0.5in}
\begin{abstract}

We present some variations affecting the association measure and
thresholding on a technique for learning Selectional Restrictions from
on-line corpora. It uses a wide-coverage noun taxonomy and a
statistical measure to generalize the appropriate semantic
classes. Evaluation measures for the Selectional Restrictions learning
task are discussed. Finally, an experimental evaluation of these
variations is reported.\\
\\
{{\bf Subject Areas}: corpus-based language modeling, computational
lexicography}
\end{abstract}

\bibliographystyle{acl}

\section{Introduction}
\label{introduction}
In recent years there has been a common agreement in the NLP research
community on the importance of having an extensive coverage of
selectional restrictions (SRs) tuned to the domain to work with.  SRs
can be seen as semantic type constraints that a word sense imposes
on the words with which it combines in the process of semantic
interpretation.  SRs may have different applications in NLP, specifically,
they may help a parser with Word Sense Selection (WSS, as in
\cite{hirst-87}), with
preferring certain structures out of several grammatical ones
\cite{whittemore-90} and finally with deciding the semantic role
played by a syntactic complement \cite{basili-et-al-92a}.
Lexicography is also interested in the acquisition of SRs (both defining in
context approach and lexical semantics work
\cite{levin-92}).

The aim of our work is to explore the feasibility of using an
statistical method for extracting SRs from on-line
corpora. Resnik \shortcite{resnik-92} developed a method for automatically
extracting class-based SRs from on-line corpora.  Ribas \shortcite{ribas-94}
performed some experiments using this basic technique and drew up some
 limitations from the corresponding results.

In this paper we will describe some substantial modifications to the
basic technique and will report the corresponding experimental
evaluation. The outline of the paper is as follows: in section
\ref{method} we summarize the basic methodology used in
\cite{ribas-94}, analyzing its limitations; in section
\ref{statistical-measures} we explore some alternative statistical measures for
ranking the hypothesized SRs; in section
\ref{evaluation} we propose some evaluation measures on the SRs-learning
problem, and use them to test the experimental results obtained
by the different techniques; finally, in section \ref{conclusions}
we draw up the final conclusions and establish future lines of research.

\section{The basic technique for learning SRs}
\label{method}

\subsection{Description}

The technique functionality can be summarized as:

\begin{description}

\item[Input] The training set, i.e. a list of complement co-occurrence triples,
{\it (verb-lemma,
syntactic-relationship, noun-lemma)} extracted from the corpus.

\item[Previous knowledge used]

A semantic
hierarchy (WordNet\footnote{WordNet is a broad-coverage lexical
database, see \cite{miller-et-al-91}.}) where words are clustered in semantic
classes,
and semantic classes are organized hierarchically. Polysemous words are
represented as instances of different classes.

\item[Output]

A set of syntactic SRs, {\it (verb-lemma, syntactic-relationship,
semantic-class, weight)}.  The final SRs must be mutually disjoint. SRs are
weighted according to the statistical evidence found in the corpus.

\item[Learning process] 3 stages:

\begin{enumerate}

\item Creation of the space of candidate classes.
\item Evaluation of the appropriateness of the candidates by
means of a statistical measure.
\item Selection of the most appropriate subset in the
candidate space to convey the SRs.
\end{enumerate}

\end {description}

\begin{table}
\begin{tabular}{||l|c|r|l||} \hline
{\it Acquired SR } & {\it Type} & {\it Assoc} &
{\it Examples of nouns in Treebank}\\ \hline
$<\! suit, suing  \!>$ & Senses & 0.41 &  suit \\
$<\! suit\_of\_clothes  \!>$ & Senses & 0.41 & suit \\
$<\! suit   \!>$ & Senses & 0.40 & suit \\
$<\! group  \!>$ & $\Uparrow$Abs & 0.35 & administration, agency, bank,
... \\
$<\! legal\_action \!>$ & Ok & 0.28 & suit \\
$<\! person, individual \!>$ & Ok & 0.23 & advocate, buyer,carrier,
client, ... \\
$<\! radical  \!>$ & Senses & 0.16 & group \\
$<\! city  \!>$ & Senses & 0.15 & proper\_name \\
$<\! admin.\_district \!>$ & Senses & 0.14 & proper\_name \\
$<\! social\_control   \!>$ & Senses & 0.11 & administration,government \\

$<\! status  \!>$ & Senses & 0.087 & government, leadership\\
$<\! activity  \!>$ & Senses & -0.01 & administration, leadership,
provision \\
$<\! cognition \!>$ & Senses & -0.04 & concern, leadership, provision,
science \\ \hline
\end{tabular}

\caption{SRs acquired for the subject of {\it seek} }
\label{seek-1-table}
\end{table}

The appropriateness of a class for expressing SRs (stage 2) is
quantified from the strength of co-occurrence of verbs and classes of
nouns in the corpus \cite{resnik-92}.  Given the verb $v$, the
syntactic-relationship $s$ and the candidate class $c$, the
Association Score, $Assoc$, between $v$ and $c$ in $s$ is defined:

\begin{eqnarray}
\label{eq-assoc-score}
Assoc(v,s,c) & = &  p(c|v,s) I(v;c|s) \nonumber \\
& = & p(c|v,s) \log \frac{p(c|v,s)}{p(c|s) \nonumber}
\end{eqnarray}

The two terms of $Assoc$ try to capture different properties:

\begin{enumerate}

\item Mutual information ratio, $I(v;c|s)$, measures the strength of the
statistical association between the given verb $v$ and the candidate
class $c$ in the given syntactic position $s$.  It compares the prior
distribution, $p(c|s)$, with the posterior distribution, $p(c|v,s)$.

\item $p(c|v,s)$ scales up the strength of the association by the
frequency of the relationship.

\end{enumerate}

\vspace*{7cm}

Probabilities are estimated by Maximum Likelihood Estimation, counting
the relative frequency of events in the corpus\footnote{Utility of
smoothing techniques on class-based distributions is dubious
\cite{resnik-93phd}.}. However, it is not obvious how to calculate
class frequencies when the training corpus is not semantically tagged
as is the case. Nevertheless, we take a simplistic approach and
calculate them in the following manner:

\begin{equation}
freq(v,s,c) = \sum_{n \in c} freq(v,s,n) \times w
\label{estimation-class-probs-crude}
\end{equation}

Where $w$  is a constant factor used to normalize the
probabilities\footnote{Resnik \shortcite{resnik-92} and Ribas
\shortcite{ribas-94} used equation
\ref{estimation-class-probs-crude}
without introducing normalization. Therefore, the estimated function
didn't accomplish probability axioms. Nevertheless, their results
should be equivalent (for our purposes) to those introducing
normalization because it shouldn't affect the relative ordering of
{\it Assoc} among rival candidate classes for the same $(v,s)$.}

\begin{equation}
\label{global-weight}
w =
\frac {\sum_{v \in \cal{V}} \sum_{s \in \cal{S}} \sum_{n \in \cal{N}}
freq(v,s,n)}
      {\sum_{v \in \cal{V}} \sum_{s \in \cal{S}} \sum_{n \in \cal{N}}
freq(v,s,n) |senses(n)|}
\end{equation}

When creating the space of candidate classes (learning process, stage
1), we use a {\it thresholding} technique to ignore as much as possible
the noise introduced in the training set.  Specifically, we consider
only those classes that have a higher number of occurrences than the
threshold. The selection of the most appropriate classes (stage 3) is
based on a global search through the candidates, in such a way that
the final classes are mutually disjoint (not related by hyperonymy).

\subsection{Evaluation}
\label{performance}

Ribas \shortcite{ribas-94} reported experimental results obtained from
the application of the above technique to learn SRs. He performed an
evaluation of the SRs obtained from a training set of 870,000 words of
the Wall Street Journal.  In this section we summarize the results and
conclusions reached in that paper.

For instance, table \ref{seek-1-table} shows the SRs acquired for the
{\it subject} position of the verb {\it seek}. {\it Type} indicates a
manual diagnosis about the class appropriateness (Ok: correct;
$\Uparrow$Abs: over-generalization; Senses: due to erroneous
senses). {\it Assoc} corresponds to the association score (higher
values appear first). Most of the induced classes are due to incorrect
senses. Thus, although $suit$ was used in the WSJ articles only in the
sense of $<\! legal\_action \!>$, the algorithm not only considered
the other senses as well ($<\! suit, suing
\!>$,$<\! suit\_of\_clothes  \!>$, $<\! suit   \!>$) , but the $Assoc$ score
ranked
them higher than the appropriate sense. We can also notice that the
$\Uparrow$Abs class, $<\! group \!>$, seems too general for the
example nouns, while one of its daughters, $<\!  people \!>$ seems to
fit the data much better.

Analyzing the results obtained from different experimental evaluation
methods, Ribas \shortcite{ribas-94} drew up some conclusions:

\begin{description}

\item[{\sl a}.] The technique achieves a good coverage.

\item[{\sl b}.] Most of the classes acquired result from the
accumulation of incorrect senses.

\item[{\sl c}.] No clear co-relation between $Assoc$ and the manual
diagnosis is found.

\item[{\sl d}.] A slight tendency to over-generalization exists
due to incorrect senses.

\end{description}

Although the performance of the presented technique seems to be quite
good, we think that some of the detected flaws could possibly be
addressed. Noise due to polysemy of the nouns involved seems to be the
main obstacle for the practicality of the technique. It makes the
association score prefer incorrect classes and jump on
over-generalizations. In this paper we are interested in exploring
various ways to make the technique more robust to noise, namely, (a)
to experiment with variations of the association score, (b) to
experiment with thresholding.

\section{Variations on the association statistical measure}
\label{statistical-measures}

 In this section we consider different variations on the association
score in order to make it more robust. The different
techniques are experimentally evaluated in section
\ref{experimental-results}.

\subsection{Variations on the prior probability}
\label{assoc-score-i-prior}

When considering the prior probability, the more independent of the
context it is the better to measure actual associations. A sensible
modification of the measure would be to consider $p(c)$ as the prior
distribution:

\[
\label{eq-assoc-score-i-prior-any-s}
Assoc'(v,s,c) = p(c|v,s) \log \frac{p(c|v,s)}{p(c)}
\]

Using the chain rule on mutual information
\cite[p. 22]{cover-thomas-91} we can mathematically relate the
different versions of $Assoc$,

\[
\label{chain-rule-assoc-2}
Assoc'(v,s,c) = p(c|v,s) \log \frac{p(c|s)}{p(c)} + Assoc(v,s,c)
\]

The first advantage of $Assoc'$ would come from this (information
theoretical) relationship. Specifically, the $Assoc'$ takes into
account the preference (selection) of syntactic positions for
particular classes. In intuitive terms, typical subjects
(e.g. $<$person, individual, ...$>$) would be preferred (to atypical
subjects as $<$suit\_of\_clothes$>$) as SRs on the subject in contrast
to $Assoc$. The second advantage is that as long as the prior
probabilities, $p(c)$, involve simpler events than those used in
$Assoc$, $p(c|s)$, the estimation is easier and more accurate
(ameliorating data sparseness).

A subsequent modification would be to
estimate the prior, $p(c)$, from the counts of all the nouns appearing
in the corpus independently of their syntactic positions (not
restricted to be heads of verbal complements).  In this
way, the estimation of $p(c)$ would be easier and more accurate.

\subsection{Estimating class probabilities from noun frequencies}
\label{estimating-class-probs-from-nouns}

In the global weighting technique presented in equation
\ref{global-weight} very polysemous nouns provide the same amount of
evidence to every sense as non-ambiguous nouns do --while less
ambiguous nouns could be more informative about the correct classes as
long as they do not carry ambiguity.

The weight introduced in (\ref{estimation-class-probs-crude}) could
alternatively be found in a local manner, in such a way that more
polysemous nouns would give less evidence to each one of their senses
than less ambiguous ones. Local weight could be obtained using
$p(c|n)$. Nevertheless, a good estimation of this probability seems
quite problematic because of the lack of tagged training material. In
absence of a better estimator we use a rather poor one as the uniform
distribution,

\[
\label{weight-sharing-senses}
w(n,c) = \tilde{p}(c|n) = \frac{|senses(n) \in c|}{|senses(n)|}
\]

Resnik \shortcite{resnik-93phd} also uses a local normalization
technique but he normalizes by the total number of classes in the
hierarchy. This scheme seems to present two problematic features (see
\cite{ribas-wp94} for more details). First, it doesn't take
dependency relationships introduced by hyperonymy into
account. Second, nouns categorized in lower levels in the taxonomy
provide less weight to each class than higher nouns.

\subsection{Other statistical measures to score SRs}
\label{variation-assoc-measure}

In this section we propose the application of other measures apart
from $Assoc$ for learning SRs: log-likelihood ratio
\cite{dunning-93}, relative entropy
\cite{cover-thomas-91}, mutual information ratio
\cite{church-hanks-90}, $\phi^{2}$ \cite{gale-church-91}.
In section (\ref{evaluation}) their experimental evaluation is
presented.

The statistical measures used to detect associations on the
distribution defined by two random variables X and Y work by measuring
the deviation of the conditional distribution, $P(X|Y)$, from the
expected distribution if both variables were considered independent,
i.e. the marginal distribution, $P(X)$.  If $P(X)$ is a good
approximation of $P(X|Y)$, association measures should be low (near
zero), otherwise deviating significantly from zero.

Table \ref{cross-table} shows the cross-table formed by
the conditional and marginal distributions in the case of $X = \{ c,
\neg c \}$ and $Y= \{ v\_s, \neg v\_s \}$. Different association
measures use the information provided in the cross-table to different
extents. Thus, $Assoc$ and mutual information ratio consider only the
deviation of the conditional probability $p(c|v,s)$ from the
corresponding marginal, $p(c)$.

\begin{table}
\centering
\begin{tabular}{|l|l|l|} \hline
& $c$ & $\neg c$  \\ \hline
$v\_s$ & $p(c | v\_s)$ & $p(\neg c | v\_s)$  \\ \hline
$\neg v\_s$ & $p(c | \neg v\_s)$ & $p( \neg c | \neg v\_s)$  \\ \hline
& $p(c)$ & $p(\neg c)$  \\ \hline
\end{tabular}

\caption{Conditional and marginal distributions}
\label{cross-table}
\end{table}

On the other hand, {\it log-likelihood} ratio and $\phi^{2}$ measure
the association between $v\_s$ and $c$ considering the deviation of
the four conditional cells in table \ref{cross-table} from the
corresponding marginals. It is plausible that the deviation of the
cells not taken into account by $Assoc$ can help on extracting useful
SRs.

Finally, it would be interesting to only use the information related
to the selectional behavior of $v\_s$, i.e. comparing the conditional
probabilities of $c$ and $\neg c$ given $v\_s$ with the corresponding
marginals. Relative entropy, $D(P(X|v\_s)||P(X))$, could do this job.

\section{Evaluation}
\label{evaluation}
\subsection{Evaluation methods of SRs}
\label{measures}

Evaluation on NLP has been crucial to fostering research in
particular areas. Evaluation of the SR learning task would provide
grounds to compare different techniques that try to abstract SRs from
corpus using WordNet (e.g, section \ref{experimental-results}). It
would also permit measuring the utility of the SRs obtained using WordNet
in comparison with other frameworks using other kinds of
knowledge. Finally it would be a powerful tool for detecting flaws of
a particular technique (e.g,
\cite{ribas-94} analysis).

However, a related and crucial issue is which linguistic tasks are
used as a reference.  SRs are useful for both lexicography and NLP. On
the one hand, from the point of view of lexicography, the goal of
evaluation would be to measure the quality of the SRs induced, (i.e., how
well the resulting classes correspond to the nouns as they were used
in the corpus). On the other hand, from the point of view of NLP, SRs
should be evaluated on their utility (i.e., how much they help on
performing the reference task).

\subsubsection{Lexicography-oriented evaluation}
\label{evaluation-measures-lexicography}

As far as lexicography (quality) is concerned, we think the main
criteria SRs acquired from corpora should meet are: (a) correct
categorization --inferred classes should correspond to the correct
senses of the words that are being generalized--, (b) appropriate
generalization level and (c) good coverage --the majority of the noun
occurrences in the corpus should be successfully generalized by the
induced SRs.

Some of the methods we could use for assessing experimentally the
accomplishment of these criteria would be:

\begin{itemize}

\item {\bf Introspection} A lexicographer  checks if the SRs
accomplish the criteria (a) and (b) above (e.g., the manual diagnosis
in table \ref{seek-1-table}). Besides the intrinsic difficulties of
this approach, it does not seem appropriate when comparing across
different techniques for learning SRs, because of its qualitative
flavor.

\item {\bf Quantification of generalization level appropriateness}
A possible measure would be the percentage of sense occurrences
included in the induced SRs which are effectively correct (from now
on called {\it Abstraction Ratio}).  Hopefully, a technique with a
higher abstraction ratio learns classes that fit the set of
examples better. A manual assessment of the ratio confirmed this behavior, as
testing sets with a lower ratio seemed to be inducing less
$\Uparrow$Abs cases.

\item {\bf Quantification of coverage} It could be measured as the
proportion of triples whose correct sense belongs to one of the SRs.

\end{itemize}

\subsubsection{NLP evaluation tasks}

The NLP tasks where SRs utility could be evaluated are
diverse. Some of them have already been introduced in section
\ref{introduction}. In the recent literature
(\cite{grishman-sterling-92}, \cite{resnik-93phd}, ...) several task
oriented schemes to test Selectional Restrictions (mainly on syntactic
ambiguity resolution) have been proposed. However, we have tested SRs
on a WSS task,
\label{wss-description} using the following scheme.  For every triple
in the testing set the algorithm selects as most appropriate that
noun-sense that has as hyperonym the SR class with highest association
score. When more than one sense belongs to the highest SR, a random
selection is performed. When no SR has been acquired, the algorithm
remains undecided.  The results of this WSS procedure are checked
against a testing-sample manually analyzed, and precision and recall
ratios are calculated. Precision is calculated as the ratio of
manual-automatic matches / number of noun occurrences disambiguated by
the procedure. Recall is computed as the ratio of manual-automatic
matches / total number of noun occurrences.

\subsection{Experimental results}
\label{experimental-results}

In order to evaluate the different variants on the association score
and the impact of thresholding we performed several experiments. In
this section we analyze the results. As training set we used the
870,000 words of WSJ material provided in the ACL/DCI version of
the Penn Treebank. The testing set consisted of 2,658 triples
corresponding to four average common verbs in the Treebank: {\it
rise}, {\it report}, {\it seek} and {\it present}. We only considered
those triples that had been correctly extracted from the Treebank and
whose noun had the correct sense included in WordNet (2,165 triples
out of the 2,658, from now on, called the {\it testing-sample}).

As evaluation measures we used coverage, abstraction ratio, and recall
and precision ratios on the WSS task (section \ref{measures}). In
addition we performed some evaluation by hand comparing the SRs
acquired by the different techniques.

\subsubsection{Comparing different techniques}

Coverage for the different techniques is
shown in table
\ref{coverage-table}. The higher the coverage, the better the technique
succeeds in
correctly generalizing more of the input examples. The labels used for
referring to the different techniques are as follows: ``$Assoc$ \&
$p(c|s)$'' corresponds to the basic association measure (section
\ref{method}), ``$Assoc$ \& Head-nouns'' and
``$Assoc$ \& All nouns'' to the techniques introduced in section
\ref{assoc-score-i-prior},
``$Assoc$ \& Normalizing'' to the local normalization
(section \ref{estimating-class-probs-from-nouns}), and finally,
{\it log-likelihood}, $D$ (relative entropy) and $I$ (mutual information
ratio) to the techniques discussed in section
\ref{variation-assoc-measure}.

\begin{table}
\centering
\begin{tabular}{||l|l|l||} \hline
Technique & Coverage (\%) \\ \hline
$Assoc$ \& All nouns & 95.7 \\
$Assoc$ \& $p(c|s)$ & 95.5 \\
$Assoc$ \& Head-nouns & 95.3 \\
$D$ & 93.7 \\
$log-likelihood$ & 92.9 \\
$Assoc$ \& Normalizing & 92.7 \\
$\phi^{2}$ & 88.2 \\
$I$ & 74.1 \\ \hline
\end{tabular}\\

\caption{Coverage Ratio}
\label{coverage-table}
\end{table}

The abstraction ratio  for the different
techniques is shown in table \ref{abstraction-table}.  In principle,
the higher abstraction ratio, the better the technique succeeds in
filtering out incorrect senses (less $\Uparrow$Abs).

\begin{table}
\centering
\begin{tabular}{||l|l||} \hline
Technique & Abs Ratio (\%) \\ \hline
$I$ & 66.6  \\
$log-likelihood$ & 64.6  \\
$\phi^{2}$ & 64.4  \\
$Assoc$ \& All nouns & 64.3  \\
$Assoc$ \& Head-nouns & 63.9  \\
$Assoc$ \& $p(c|s)$ & 63  \\
$D$ & 62.3  \\
$Assoc$ \& Normalizing & 58.5  \\ \hline
\end{tabular}\\

\caption{Abstraction Ratio}
\label{abstraction-table}
\end{table}

The precision and recall ratios on the noun WSS task for the different
techniques are shown in table
\ref{precision-table}.  In principle, the higher
the precision and recall ratios the better the technique succeeds in
inducing appropriate SRs for the disambiguation task.

\begin{table}
\centering
\begin{tabular}{||l|l|l||} \hline
Technique & Prec. (\%) & Rec. (\%) \\ \hline
$Assoc$ \& All nouns & 80.3 & 78.5 \\
$Assoc$ \& $p(c|s)$ & 79.9 & 77.9 \\
$Assoc$ \& Head-nouns & 78.5 & 76.7  \\
$log-likelihood$ & 77.2 & 74.4 \\
$D$ & 75.9 & 74.1 \\
$Assoc$ \& Normalizing & 75.9 & 73.3 \\
$\phi^{2}$ & 67.8 & 63 \\
$I$ & 50.4 & 45.7 \\ \hline
Guessing Heuristic& 62.7 & 62.7 \\ \hline
\end{tabular}\\

\caption{Precision and Recall on the WSS task}
\label{precision-table}
\end{table}

As far as the evaluation measures try to account for different
phenomena the goodness of a particular technique should be quantified
as a trade-off. Most of the results are very similar (differences are
not statistically significative). Therefore we should be cautious when
extrapolating the results. Some of the conclusions from the tables
above are:

\begin{enumerate}

\item $\phi^{2}$ and $I$ get sensibly worse results than other
measures (although abstraction is quite good).

\item The local normalizing technique using the
uniform distribution does not help. It seems that by using the local
weighting we misinform the algorithm. The problem is the reduced
weight that polysemous nouns get, while they seem to be the most
informative\footnote{In some way, it conforms to Zipf-law
\cite{zipf-45}: noun frequency and polysemy are correlated.}. However,
a better informed kind of local weight (section
\ref{conclusions}) should improve the technique significantly.

\item All versions of $Assoc$ (except the local normalization) get
good results. Specially the two techniques that exploit a simpler
prior distribution, which seem to improve the basic technique.

\item {\it log-likelihood} and $D$  seem to get slightly worse
results than $Assoc$ techniques, although the results are very similar.

\end{enumerate}

\subsubsection{Thresholding}

\begin{figure}

%\include{/usr/usuaris/doctorat/ribas/perl/doc/grafics/prec-rec-covs-covw-abs-1-20-one-column}

% GNUPLOT: LaTeX picture
\setlength{\unitlength}{0.240900pt}
\ifx\plotpoint\undefined\newsavebox{\plotpoint}\fi
\sbox{\plotpoint}{\rule[-0.200pt]{0.400pt}{0.400pt}}%
\begin{picture}(900,720)(0,0)
\font\gnuplot=cmr10 at 10pt
\gnuplot
\sbox{\plotpoint}{\rule[-0.200pt]{0.400pt}{0.400pt}}%
\put(220.0,113.0){\rule[-0.200pt]{0.400pt}{140.686pt}}
\put(220.0,129.0){\rule[-0.200pt]{4.818pt}{0.400pt}}
\put(198,129){\makebox(0,0)[r]{60}}
\put(816.0,129.0){\rule[-0.200pt]{4.818pt}{0.400pt}}
\put(220.0,170.0){\rule[-0.200pt]{4.818pt}{0.400pt}}
\put(198,170){\makebox(0,0)[r]{65}}
\put(816.0,170.0){\rule[-0.200pt]{4.818pt}{0.400pt}}
\put(220.0,210.0){\rule[-0.200pt]{4.818pt}{0.400pt}}
\put(198,210){\makebox(0,0)[r]{70}}
\put(816.0,210.0){\rule[-0.200pt]{4.818pt}{0.400pt}}
\put(220.0,251.0){\rule[-0.200pt]{4.818pt}{0.400pt}}
\put(198,251){\makebox(0,0)[r]{75}}
\put(816.0,251.0){\rule[-0.200pt]{4.818pt}{0.400pt}}
\put(220.0,291.0){\rule[-0.200pt]{4.818pt}{0.400pt}}
\put(198,291){\makebox(0,0)[r]{80}}
\put(816.0,291.0){\rule[-0.200pt]{4.818pt}{0.400pt}}
\put(220.0,332.0){\rule[-0.200pt]{4.818pt}{0.400pt}}
\put(198,332){\makebox(0,0)[r]{85}}
\put(816.0,332.0){\rule[-0.200pt]{4.818pt}{0.400pt}}
\put(220.0,373.0){\rule[-0.200pt]{4.818pt}{0.400pt}}
\put(198,373){\makebox(0,0)[r]{90}}
\put(816.0,373.0){\rule[-0.200pt]{4.818pt}{0.400pt}}
\put(220.0,413.0){\rule[-0.200pt]{4.818pt}{0.400pt}}
\put(198,413){\makebox(0,0)[r]{95}}
\put(816.0,413.0){\rule[-0.200pt]{4.818pt}{0.400pt}}
\put(220.0,454.0){\rule[-0.200pt]{4.818pt}{0.400pt}}
\put(198,454){\makebox(0,0)[r]{100}}
\put(816.0,454.0){\rule[-0.200pt]{4.818pt}{0.400pt}}
\put(220.0,113.0){\rule[-0.200pt]{0.400pt}{4.818pt}}
\put(220,68){\makebox(0,0){0}}
\put(220.0,677.0){\rule[-0.200pt]{0.400pt}{4.818pt}}
\put(374.0,113.0){\rule[-0.200pt]{0.400pt}{4.818pt}}
\put(374,68){\makebox(0,0){5}}
\put(374.0,677.0){\rule[-0.200pt]{0.400pt}{4.818pt}}
\put(528.0,113.0){\rule[-0.200pt]{0.400pt}{4.818pt}}
\put(528,68){\makebox(0,0){10}}
\put(528.0,677.0){\rule[-0.200pt]{0.400pt}{4.818pt}}
\put(682.0,113.0){\rule[-0.200pt]{0.400pt}{4.818pt}}
\put(682,68){\makebox(0,0){15}}
\put(682.0,677.0){\rule[-0.200pt]{0.400pt}{4.818pt}}
\put(836.0,113.0){\rule[-0.200pt]{0.400pt}{4.818pt}}
\put(836,68){\makebox(0,0){20}}
\put(836.0,677.0){\rule[-0.200pt]{0.400pt}{4.818pt}}
\put(220.0,113.0){\rule[-0.200pt]{148.394pt}{0.400pt}}
\put(836.0,113.0){\rule[-0.200pt]{0.400pt}{140.686pt}}
\put(220.0,697.0){\rule[-0.200pt]{148.394pt}{0.400pt}}
\put(45,405){\makebox(0,0){\%}}
\put(528,23){\makebox(0,0){Threshold}}
\put(220.0,113.0){\rule[-0.200pt]{0.400pt}{140.686pt}}
\put(706,632){\makebox(0,0)[r]{Precision}}
\put(728.0,632.0){\rule[-0.200pt]{15.899pt}{0.400pt}}
\put(251,276){\usebox{\plotpoint}}
\multiput(251.00,276.59)(3.382,0.477){7}{\rule{2.580pt}{0.115pt}}
\multiput(251.00,275.17)(25.645,5.000){2}{\rule{1.290pt}{0.400pt}}
\multiput(282.00,281.59)(3.270,0.477){7}{\rule{2.500pt}{0.115pt}}
\multiput(282.00,280.17)(24.811,5.000){2}{\rule{1.250pt}{0.400pt}}
\multiput(312.00,286.61)(6.714,0.447){3}{\rule{4.233pt}{0.108pt}}
\multiput(312.00,285.17)(22.214,3.000){2}{\rule{2.117pt}{0.400pt}}
\put(343,288.67){\rule{7.468pt}{0.400pt}}
\multiput(343.00,288.17)(15.500,1.000){2}{\rule{3.734pt}{0.400pt}}
\put(405,288.17){\rule{6.300pt}{0.400pt}}
\multiput(405.00,289.17)(17.924,-2.000){2}{\rule{3.150pt}{0.400pt}}
\put(436,287.67){\rule{7.227pt}{0.400pt}}
\multiput(436.00,287.17)(15.000,1.000){2}{\rule{3.613pt}{0.400pt}}
\put(466,289.17){\rule{6.300pt}{0.400pt}}
\multiput(466.00,288.17)(17.924,2.000){2}{\rule{3.150pt}{0.400pt}}
\put(374.0,290.0){\rule[-0.200pt]{7.468pt}{0.400pt}}
\put(528,291.17){\rule{6.300pt}{0.400pt}}
\multiput(528.00,290.17)(17.924,2.000){2}{\rule{3.150pt}{0.400pt}}
\put(559,291.67){\rule{7.468pt}{0.400pt}}
\multiput(559.00,292.17)(15.500,-1.000){2}{\rule{3.734pt}{0.400pt}}
\put(590,292.17){\rule{6.100pt}{0.400pt}}
\multiput(590.00,291.17)(17.339,2.000){2}{\rule{3.050pt}{0.400pt}}
\put(620,294.17){\rule{6.300pt}{0.400pt}}
\multiput(620.00,293.17)(17.924,2.000){2}{\rule{3.150pt}{0.400pt}}
\put(497.0,291.0){\rule[-0.200pt]{7.468pt}{0.400pt}}
\multiput(682.00,294.93)(3.382,-0.477){7}{\rule{2.580pt}{0.115pt}}
\multiput(682.00,295.17)(25.645,-5.000){2}{\rule{1.290pt}{0.400pt}}
\put(713,289.67){\rule{7.468pt}{0.400pt}}
\multiput(713.00,290.17)(15.500,-1.000){2}{\rule{3.734pt}{0.400pt}}
\put(744,288.67){\rule{7.227pt}{0.400pt}}
\multiput(744.00,289.17)(15.000,-1.000){2}{\rule{3.613pt}{0.400pt}}
\put(774,287.67){\rule{7.468pt}{0.400pt}}
\multiput(774.00,288.17)(15.500,-1.000){2}{\rule{3.734pt}{0.400pt}}
\put(805,288.17){\rule{6.300pt}{0.400pt}}
\multiput(805.00,287.17)(17.924,2.000){2}{\rule{3.150pt}{0.400pt}}
\put(750,632){\raisebox{-.8pt}{\makebox(0,0){$\Diamond$}}}
\put(251,276){\raisebox{-.8pt}{\makebox(0,0){$\Diamond$}}}
\put(282,281){\raisebox{-.8pt}{\makebox(0,0){$\Diamond$}}}
\put(312,286){\raisebox{-.8pt}{\makebox(0,0){$\Diamond$}}}
\put(343,289){\raisebox{-.8pt}{\makebox(0,0){$\Diamond$}}}
\put(374,290){\raisebox{-.8pt}{\makebox(0,0){$\Diamond$}}}
\put(405,290){\raisebox{-.8pt}{\makebox(0,0){$\Diamond$}}}
\put(436,288){\raisebox{-.8pt}{\makebox(0,0){$\Diamond$}}}
\put(466,289){\raisebox{-.8pt}{\makebox(0,0){$\Diamond$}}}
\put(497,291){\raisebox{-.8pt}{\makebox(0,0){$\Diamond$}}}
\put(528,291){\raisebox{-.8pt}{\makebox(0,0){$\Diamond$}}}
\put(559,293){\raisebox{-.8pt}{\makebox(0,0){$\Diamond$}}}
\put(590,292){\raisebox{-.8pt}{\makebox(0,0){$\Diamond$}}}
\put(620,294){\raisebox{-.8pt}{\makebox(0,0){$\Diamond$}}}
\put(651,296){\raisebox{-.8pt}{\makebox(0,0){$\Diamond$}}}
\put(682,296){\raisebox{-.8pt}{\makebox(0,0){$\Diamond$}}}
\put(713,291){\raisebox{-.8pt}{\makebox(0,0){$\Diamond$}}}
\put(744,290){\raisebox{-.8pt}{\makebox(0,0){$\Diamond$}}}
\put(774,289){\raisebox{-.8pt}{\makebox(0,0){$\Diamond$}}}
\put(805,288){\raisebox{-.8pt}{\makebox(0,0){$\Diamond$}}}
\put(836,290){\raisebox{-.8pt}{\makebox(0,0){$\Diamond$}}}
\put(651.0,296.0){\rule[-0.200pt]{7.468pt}{0.400pt}}
\put(706,587){\makebox(0,0)[r]{Recall}}
\multiput(728,587)(20.756,0.000){4}{\usebox{\plotpoint}}
\put(794,587){\usebox{\plotpoint}}
\put(251,275){\usebox{\plotpoint}}
\multiput(251,275)(20.745,-0.669){2}{\usebox{\plotpoint}}
\put(292.44,275.04){\usebox{\plotpoint}}
\multiput(312,277)(20.712,-1.336){2}{\usebox{\plotpoint}}
\put(354.54,274.63){\usebox{\plotpoint}}
\multiput(374,274)(20.491,-3.305){2}{\usebox{\plotpoint}}
\put(416.37,269.73){\usebox{\plotpoint}}
\multiput(436,271)(20.473,-3.412){2}{\usebox{\plotpoint}}
\put(478.13,264.83){\usebox{\plotpoint}}
\multiput(497,263)(20.756,0.000){2}{\usebox{\plotpoint}}
\put(540.28,262.21){\usebox{\plotpoint}}
\multiput(559,261)(20.745,-0.669){2}{\usebox{\plotpoint}}
\put(602.49,259.58){\usebox{\plotpoint}}
\multiput(620,259)(20.745,-0.669){2}{\usebox{\plotpoint}}
\put(664.73,258.00){\usebox{\plotpoint}}
\multiput(682,258)(20.491,-3.305){2}{\usebox{\plotpoint}}
\put(726.42,250.84){\usebox{\plotpoint}}
\multiput(744,248)(20.756,0.000){2}{\usebox{\plotpoint}}
\put(788.45,247.53){\usebox{\plotpoint}}
\multiput(805,247)(20.756,0.000){2}{\usebox{\plotpoint}}
\put(836,247){\usebox{\plotpoint}}
\put(750,587){\makebox(0,0){$+$}}
\put(251,275){\makebox(0,0){$+$}}
\put(282,274){\makebox(0,0){$+$}}
\put(312,277){\makebox(0,0){$+$}}
\put(343,275){\makebox(0,0){$+$}}
\put(374,274){\makebox(0,0){$+$}}
\put(405,269){\makebox(0,0){$+$}}
\put(436,271){\makebox(0,0){$+$}}
\put(466,266){\makebox(0,0){$+$}}
\put(497,263){\makebox(0,0){$+$}}
\put(528,263){\makebox(0,0){$+$}}
\put(559,261){\makebox(0,0){$+$}}
\put(590,260){\makebox(0,0){$+$}}
\put(620,259){\makebox(0,0){$+$}}
\put(651,258){\makebox(0,0){$+$}}
\put(682,258){\makebox(0,0){$+$}}
\put(713,253){\makebox(0,0){$+$}}
\put(744,248){\makebox(0,0){$+$}}
\put(774,248){\makebox(0,0){$+$}}
\put(805,247){\makebox(0,0){$+$}}
\put(836,247){\makebox(0,0){$+$}}
\sbox{\plotpoint}{\rule[-0.400pt]{0.800pt}{0.800pt}}%
\put(706,542){\makebox(0,0)[r]{Coverage}}
\put(728.0,542.0){\rule[-0.400pt]{15.899pt}{0.800pt}}
\put(251,442){\usebox{\plotpoint}}
\multiput(251.00,440.09)(0.989,-0.507){25}{\rule{1.750pt}{0.122pt}}
\multiput(251.00,440.34)(27.368,-16.000){2}{\rule{0.875pt}{0.800pt}}
\multiput(282.00,424.06)(4.622,-0.560){3}{\rule{5.000pt}{0.135pt}}
\multiput(282.00,424.34)(19.622,-5.000){2}{\rule{2.500pt}{0.800pt}}
\put(312,417.34){\rule{6.400pt}{0.800pt}}
\multiput(312.00,419.34)(17.716,-4.000){2}{\rule{3.200pt}{0.800pt}}
\multiput(374.00,415.08)(1.656,-0.514){13}{\rule{2.680pt}{0.124pt}}
\multiput(374.00,415.34)(25.438,-10.000){2}{\rule{1.340pt}{0.800pt}}
\put(405,406.34){\rule{7.468pt}{0.800pt}}
\multiput(405.00,405.34)(15.500,2.000){2}{\rule{3.734pt}{0.800pt}}
\multiput(436.00,407.08)(2.090,-0.520){9}{\rule{3.200pt}{0.125pt}}
\multiput(436.00,407.34)(23.358,-8.000){2}{\rule{1.600pt}{0.800pt}}
\multiput(466.00,399.07)(3.253,-0.536){5}{\rule{4.333pt}{0.129pt}}
\multiput(466.00,399.34)(22.006,-6.000){2}{\rule{2.167pt}{0.800pt}}
\put(497,393.84){\rule{7.468pt}{0.800pt}}
\multiput(497.00,393.34)(15.500,1.000){2}{\rule{3.734pt}{0.800pt}}
\put(528,392.84){\rule{7.468pt}{0.800pt}}
\multiput(528.00,394.34)(15.500,-3.000){2}{\rule{3.734pt}{0.800pt}}
\put(559,389.84){\rule{7.468pt}{0.800pt}}
\multiput(559.00,391.34)(15.500,-3.000){2}{\rule{3.734pt}{0.800pt}}
\put(590,387.84){\rule{7.227pt}{0.800pt}}
\multiput(590.00,388.34)(15.000,-1.000){2}{\rule{3.613pt}{0.800pt}}
\put(620,385.34){\rule{6.400pt}{0.800pt}}
\multiput(620.00,387.34)(17.716,-4.000){2}{\rule{3.200pt}{0.800pt}}
\put(343.0,417.0){\rule[-0.400pt]{7.468pt}{0.800pt}}
\put(682,384.84){\rule{7.468pt}{0.800pt}}
\multiput(682.00,383.34)(15.500,3.000){2}{\rule{3.734pt}{0.800pt}}
\multiput(713.00,386.08)(2.577,-0.526){7}{\rule{3.743pt}{0.127pt}}
\multiput(713.00,386.34)(23.232,-7.000){2}{\rule{1.871pt}{0.800pt}}
\put(744,380.84){\rule{7.227pt}{0.800pt}}
\multiput(744.00,379.34)(15.000,3.000){2}{\rule{3.613pt}{0.800pt}}
\put(774,381.34){\rule{7.468pt}{0.800pt}}
\multiput(774.00,382.34)(15.500,-2.000){2}{\rule{3.734pt}{0.800pt}}
\multiput(805.00,380.08)(1.349,-0.511){17}{\rule{2.267pt}{0.123pt}}
\multiput(805.00,380.34)(26.295,-12.000){2}{\rule{1.133pt}{0.800pt}}
\put(750,542){\raisebox{-.8pt}{\makebox(0,0){$\Box$}}}
\put(251,442){\raisebox{-.8pt}{\makebox(0,0){$\Box$}}}
\put(282,426){\raisebox{-.8pt}{\makebox(0,0){$\Box$}}}
\put(312,421){\raisebox{-.8pt}{\makebox(0,0){$\Box$}}}
\put(343,417){\raisebox{-.8pt}{\makebox(0,0){$\Box$}}}
\put(374,417){\raisebox{-.8pt}{\makebox(0,0){$\Box$}}}
\put(405,407){\raisebox{-.8pt}{\makebox(0,0){$\Box$}}}
\put(436,409){\raisebox{-.8pt}{\makebox(0,0){$\Box$}}}
\put(466,401){\raisebox{-.8pt}{\makebox(0,0){$\Box$}}}
\put(497,395){\raisebox{-.8pt}{\makebox(0,0){$\Box$}}}
\put(528,396){\raisebox{-.8pt}{\makebox(0,0){$\Box$}}}
\put(559,393){\raisebox{-.8pt}{\makebox(0,0){$\Box$}}}
\put(590,390){\raisebox{-.8pt}{\makebox(0,0){$\Box$}}}
\put(620,389){\raisebox{-.8pt}{\makebox(0,0){$\Box$}}}
\put(651,385){\raisebox{-.8pt}{\makebox(0,0){$\Box$}}}
\put(682,385){\raisebox{-.8pt}{\makebox(0,0){$\Box$}}}
\put(713,388){\raisebox{-.8pt}{\makebox(0,0){$\Box$}}}
\put(744,381){\raisebox{-.8pt}{\makebox(0,0){$\Box$}}}
\put(774,384){\raisebox{-.8pt}{\makebox(0,0){$\Box$}}}
\put(805,382){\raisebox{-.8pt}{\makebox(0,0){$\Box$}}}
\put(836,370){\raisebox{-.8pt}{\makebox(0,0){$\Box$}}}
\put(651.0,385.0){\rule[-0.400pt]{7.468pt}{0.800pt}}
\sbox{\plotpoint}{\rule[-0.500pt]{1.000pt}{1.000pt}}%
\put(706,497){\makebox(0,0)[r]{Abstraction Ratio}}
\multiput(728,497)(20.756,0.000){4}{\usebox{\plotpoint}}
\put(794,497){\usebox{\plotpoint}}
\put(251,192){\usebox{\plotpoint}}
\multiput(251,192)(19.561,-6.941){2}{\usebox{\plotpoint}}
\multiput(282,181)(17.004,-11.902){2}{\usebox{\plotpoint}}
\put(325.48,159.13){\usebox{\plotpoint}}
\multiput(343,158)(20.585,-2.656){2}{\usebox{\plotpoint}}
\put(387.45,154.43){\usebox{\plotpoint}}
\multiput(405,155)(20.745,-0.669){2}{\usebox{\plotpoint}}
\put(449.11,150.07){\usebox{\plotpoint}}
\multiput(466,145)(20.377,-3.944){2}{\usebox{\plotpoint}}
\put(510.00,137.74){\usebox{\plotpoint}}
\multiput(528,136)(20.745,0.669){2}{\usebox{\plotpoint}}
\put(572.16,137.00){\usebox{\plotpoint}}
\multiput(590,137)(20.744,-0.691){2}{\usebox{\plotpoint}}
\put(634.29,134.16){\usebox{\plotpoint}}
\multiput(651,132)(20.659,-1.999){2}{\usebox{\plotpoint}}
\put(696.21,127.62){\usebox{\plotpoint}}
\multiput(713,126)(20.585,-2.656){2}{\usebox{\plotpoint}}
\put(758.07,120.59){\usebox{\plotpoint}}
\multiput(774,119)(20.585,-2.656){2}{\usebox{\plotpoint}}
\put(819.99,115.48){\usebox{\plotpoint}}
\put(836,116){\usebox{\plotpoint}}
\put(750,497){\makebox(0,0){$\times$}}
\put(251,192){\makebox(0,0){$\times$}}
\put(282,181){\makebox(0,0){$\times$}}
\put(312,160){\makebox(0,0){$\times$}}
\put(343,158){\makebox(0,0){$\times$}}
\put(374,154){\makebox(0,0){$\times$}}
\put(405,155){\makebox(0,0){$\times$}}
\put(436,154){\makebox(0,0){$\times$}}
\put(466,145){\makebox(0,0){$\times$}}
\put(497,139){\makebox(0,0){$\times$}}
\put(528,136){\makebox(0,0){$\times$}}
\put(559,137){\makebox(0,0){$\times$}}
\put(590,137){\makebox(0,0){$\times$}}
\put(620,136){\makebox(0,0){$\times$}}
\put(651,132){\makebox(0,0){$\times$}}
\put(682,129){\makebox(0,0){$\times$}}
\put(713,126){\makebox(0,0){$\times$}}
\put(744,122){\makebox(0,0){$\times$}}
\put(774,119){\makebox(0,0){$\times$}}
\put(805,115){\makebox(0,0){$\times$}}
\put(836,116){\makebox(0,0){$\times$}}
\end{picture}

\caption{{\it Assoc}: Evaluation ratios vs. Threshold}
\label{prec-rec-abs-cov-figure}
\end{figure}
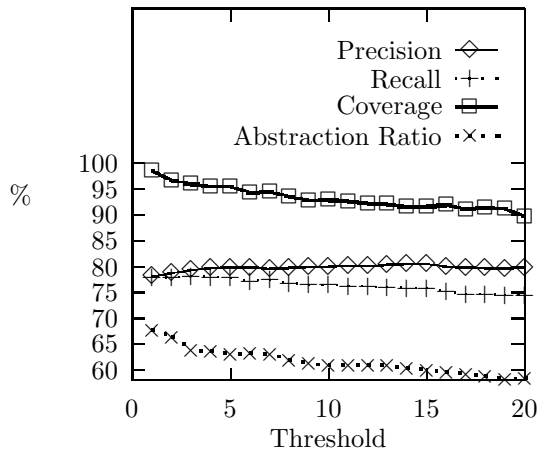

We were also interested in measuring the impact of thresholding on the
SRs acquired. In figure \ref{prec-rec-abs-cov-figure} we can see the
different evaluation measures of the basic technique when varying the
threshold. Precision and recall coincide when no candidate classes are
refused ($threshold = 1$). However, as it might be expected, as the
threshold increases (i.e. some cases are not classified) the two
ratios slightly diverge (precision increases and recall diminishes).

Figure \ref{prec-rec-abs-cov-figure} also shows the impact of
thresholding on coverage and abstraction ratios. Both decrease when
threshold increases, probably because when the rejecting threshold is
low, small classes that fit the data well can be induced, learning
over-general or incomplete SRs otherwise.

Finally, it seems that precision and abstraction ratios are in inverse
co-relation (as precision grows, abstraction decreases). In terms of
WSS, general classes may be performing better than classes that fit
the data better. Nevertheless, this relationship should be further
explored in future work.

\section{Conclusions and future work}
\label{conclusions}

In this paper we have presented some variations affecting the
association measure and thresholding on the basic technique for
learning SRs from on-line corpora.  We proposed some evaluation
measures for the SRs learning task. Finally, experimental results on
these variations were reported. We can conclude that some of these
variations seem to improve the results obtained using the basic
technique. However, although the technique still seems far from
practical application to NLP tasks, it may be most useful for providing
experimental insight to lexicographers. Future lines of research will
mainly concentrate on improving the local normalization technique by
solving the noun sense ambiguity. We have foreseen the application of
the following techniques:

\begin{itemize}

\item Simple techniques to decide the best sense $c$ given the target
	noun $n$ using estimates of the n-grams: $P(c)$, $P(c|n)$,
         $P(c|v,s)$ and $P(c|v,s,n)$, obtained from supervised and
         un-supervised corpora.

\item Combining the different n-grams by means of smoothing
         techniques.

\item Calculating $P(c|v,s,n)$ combining $P(n|c)$ and
	 $P(c|v,s)$, and applying the EM Algorithm
	 \cite{dempster-et-al-77} to improve the model.

\item Using the WordNet hierarchy as a source of backing-off
	knowledge, in such a way that if n-grams composed by $c$
	aren't enough to decide the best sense (are equal to zero),
	the tri-grams of ancestor classes could be used instead.
\end{itemize}

%\bibliography{/usr/usuaris/doctorat/ribas/perl/doc/bibliografia/biblio}

\begin{thebibliography}{}

\bibitem[\protect\citename{Basili \bgroup et al.\egroup
  }1992]{basili-et-al-92a}
R.~Basili, M.T. Pazienza, and P.~Velardi.
\newblock 1992.
\newblock Computational lexicons: the neat examples and the odd exemplars.
\newblock In {\em Procs 3rd ANLP}, Trento, Italy, April.

\bibitem[\protect\citename{Church and Hanks}1990]{church-hanks-90}
K.W. Church and P.~Hanks.
\newblock 1990.
\newblock Word association norms, mutual information and lexicography.
\newblock {\em Computational Linguistics}, 16(1).

\bibitem[\protect\citename{Cover and Thomas}1991]{cover-thomas-91}
T.M. Cover and J.A. Thomas, editors.
\newblock 1991.
\newblock {\em Elements of Information Theory}.
\newblock John Wiley.

\bibitem[\protect\citename{Dempster \bgroup et al.\egroup
  }1977]{dempster-et-al-77}
A.~P. Dempster, N.~M. Laird, and D.~B. Rubin.
\newblock 1977.
\newblock Maximum likelihood from incomplete data via the em algorithm.
\newblock {\em Journal of the Royal Statistical Society}, 39(B):1--38.

\bibitem[\protect\citename{Dunning}1993]{dunning-93}
T.~Dunning.
\newblock 1993.
\newblock Accurate methods for the statistics of surprise and coincidence.
\newblock {\em Computational Linguistics}, 19(1):61--74.

\bibitem[\protect\citename{Gale and Church}1991]{gale-church-91}
W.~Gale and K.~W. Church.
\newblock 1991.
\newblock Identifying word correspondences in parallel texts.
\newblock In {\em DARPA Speech and Natural Language Workshop}, Pacific Grove,
  California, February.
\newblock .

\bibitem[\protect\citename{Grishman and Sterling}1992]{grishman-sterling-92}
R.~Grishman and J.~Sterling.
\newblock 1992.
\newblock Acquisition of selectional patterns.
\newblock In {\em COLING}, Nantes, France, march.

\bibitem[\protect\citename{Hirst}1987]{hirst-87}
G. Hirst.
\newblock 1987.
\newblock {\em Semantic interpretation and the resolution of ambiguity}.
\newblock Cambridge University Press.

\bibitem[\protect\citename{Levin}1992]{levin-92}
B. Levin.
\newblock 1992.
\newblock {\em Towards a lexical organization of English verbs}.
\newblock University of Chicago Press.

\bibitem[\protect\citename{Miller \bgroup et al.\egroup }1991]{miller-et-al-91}
G.~Miller, R.~Beckwith, C.~Fellbaum, D.~Gross, and K.~Miller.
\newblock 1991.
\newblock Five papers on wordnet.
\newblock {\em International Journal of Lexicography}.

\bibitem[\protect\citename{Resnik}1992]{resnik-92}
P.~S. Resnik.
\newblock 1992.
\newblock Wordnet and distributional analysis: A class-based approach to
  lexical discovery.
\newblock In {\em AAAI Symposium on Probabilistic Approaches to NL}, San Jose,
  CA.

\bibitem[\protect\citename{Resnik}1993]{resnik-93phd}
P.~S. Resnik.
\newblock 1993.
\newblock {\em Selection and Information: A Class-Based Approach to lexical
  relationships}.
\newblock {Ph.D.} thesis, Computer and Information Science Department,
  University of Pennsylvania.

\bibitem[\protect\citename{Ribas}1994a]{ribas-94}
F.~Ribas.
\newblock 1994a.
\newblock An experiment on learning appropriate selectional restrictions from a
  parsed corpus.
\newblock In {\em COLING}, Kyoto, Japan, August.

\bibitem[\protect\citename{Ribas}1994b]{ribas-wp94}
F.~Ribas.
\newblock 1994b.
\newblock Learning more appropriate selectional restrictions.
\newblock Technical report, ESPRIT BRA-7315 ACQUILEX-II WP.

\bibitem[\protect\citename{Whittemore \bgroup et al.\egroup
  }1990]{whittemore-90}
G.~Whittemore, K.~Ferrara, and H.~Brunner.
\newblock 1990.
\newblock Empirical study of predictive powers of simple attachment schemes for
  post-modifier prepositional phrases.
\newblock In {\em Procs. ACL}, Pennsylvania.

\bibitem[\protect\citename{Zipf}1945]{zipf-45}
G.~K. Zipf.
\newblock 1945.
\newblock The meaning-frequency relationship of words.
\newblock {\em The Journal of General Psychology}, 33:251--256.

\end{thebibliography}

(Acquilex-II Working Papers can be obtained by sending a request to
{\tt cide@cup.cam.uk})

\end{document}